\documentclass[runningheads]{llncs}
\usepackage{graphicx}
\usepackage{graphics}
\usepackage{amsmath,amssymb} 
\usepackage{color}
\usepackage{wrapfig}
\usepackage{array}
\usepackage{multirow}
\usepackage{dsfont}
\usepackage{hyphenat}
\begin{document}
\pagestyle{headings}
\mainmatter
\def\ECCV18SubNumber{7}  

\title{Complex-YOLO: An Euler-Region-Proposal for Real-time 3D Object Detection on Point Clouds}



\titlerunning{Complex-YOLO: Real-time 3D Object Detection on Point Clouds}
\authorrunning{Simon et al.}

\author{Authors}
\author{
  Martin Simon\textsuperscript{\dag*},
  Stefan Milz\textsuperscript{\dag}, 
  Karl Amende\textsuperscript{\dag*},
  Horst-Michael Gross\textsuperscript{*}
}

\institute{Valeo Schalter und Sensoren GmbH\textsuperscript{\dag},
	Ilmenau University of Technology\textsuperscript{*}\\
	\email{ \{martin.simon,stefan.milz,karl.amende\}@valeo.com} \\ \email{ horst-michael.gross@tu-ilmenau.de}
}


\maketitle

\begin{abstract}
Lidar based 3D object detection is inevitable for autonomous driving, because it directly links to environmental understanding and therefore builds the base for prediction and motion planning. The capacity of inferencing highly sparse 3D data in real-time is an ill-posed problem for lots of other application areas besides automated vehicles, e.g. augmented reality, personal robotics or industrial automation. We introduce Complex-YOLO, a state of the art real-time 3D object detection network on point clouds only. In this work, we describe a network that expands YOLOv2, a fast 2D standard object detector for RGB images, by a specific complex regression strategy to estimate multi-class 3D boxes in Cartesian space. Thus, we propose a specific Euler-Region-Proposal Network (E-RPN) to estimate the pose of the object by adding an imaginary and a real fraction to the regression network. This ends up in a closed complex space and avoids singularities, which occur by single angle estimations. The E-RPN supports to generalize well during training. Our experiments on the KITTI benchmark suite show that we outperform current leading methods for 3D object detection specifically in terms of efficiency. We achieve state of the art results for cars, pedestrians and cyclists by being more than five times faster than the fastest competitor. Further, our model is capable of estimating all eight KITTI-classes, including Vans, Trucks or sitting pedestrians simultaneously with high accuracy.

\end{abstract}

\section{Introduction}

Point cloud processing is becoming more and more important for autonomous driving due to the strong improvement of automotive Lidar sensors in the recent years. The sensors of suppliers are capable to deliver 3D points of the surrounding environment in real-time. The advantage is a direct measurement of the distance of encompassing objects \cite{Geiger:2012:WRA:2354409.2354978}. This allows us to develop object detection algorithms for autonomous driving that estimate the position and the heading of different objects accurately in 3D \cite{DBLP:journals/corr/ChenMWLX16} \cite{DBLP:journals/corr/abs-1711-06396} \cite{DBLP:journals/corr/EngelckeRWTP16} \cite{DBLP:journals/corr/abs-1711-08488} \cite{Wang-RSS-15} \cite{ku2017joint} \cite{DBLP:journals/corr/LiZX16} \cite{DBLP:journals/corr/Li16p}. Compared to images, Lidar point clouds are sparse with a varying density distributed all over the measurement area. Those points are unordered, they interact locally and could mainly be not analyzed isolated. Point cloud processing should always be invariant to basic transformations \cite{DBLP:journals/corr/QiSMG16} \cite{DBLP:journals/corr/QiYSG17}.

\begin{figure}[!t]
\centering
\includegraphics[width=\textwidth]{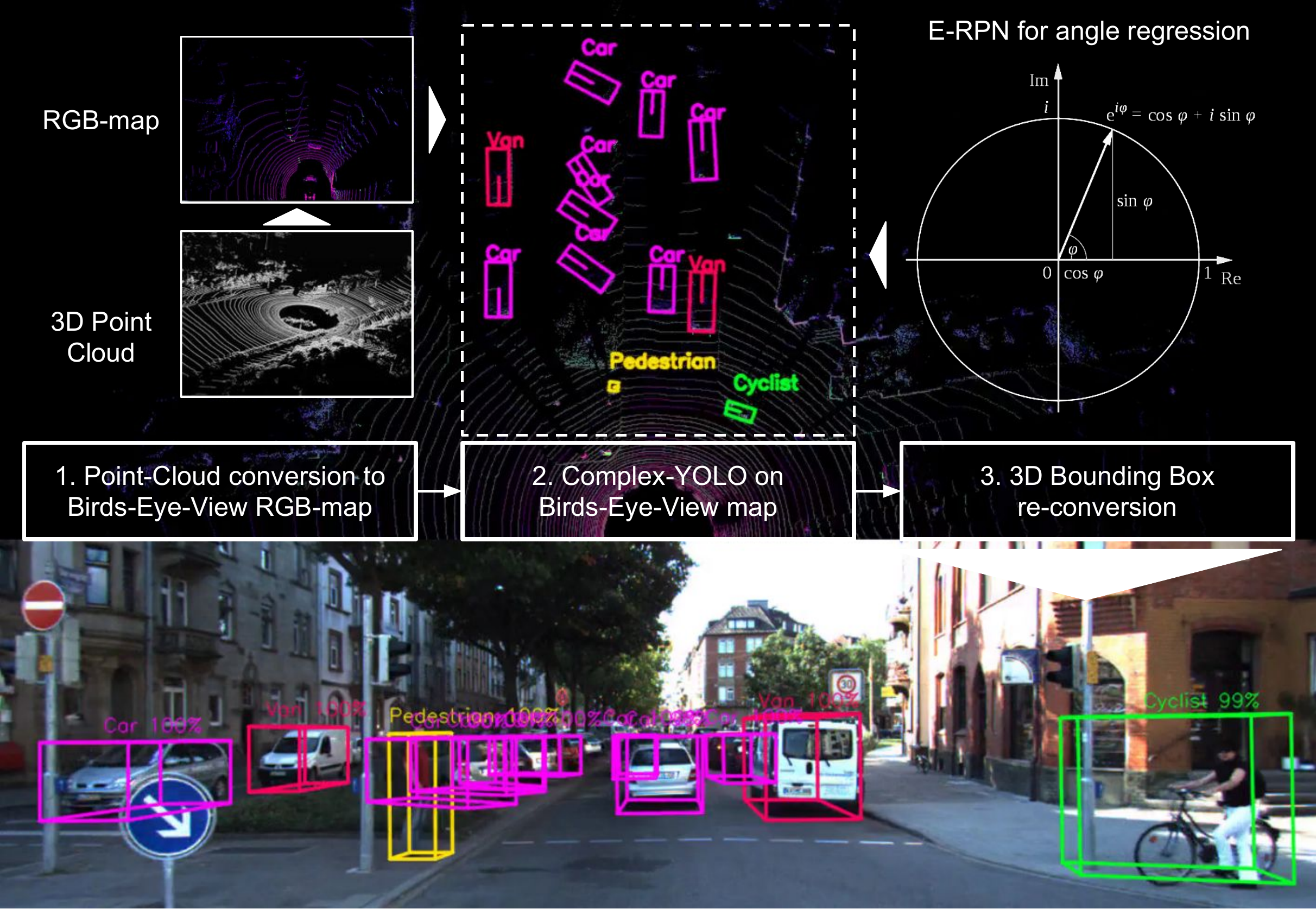}
\caption{Complex-YOLO is a very efficient model that directly operates on Lidar only based birds-eye-view RGB-maps to estimate and localize accurate 3D multiclass bounding boxes. The upper part of the figure shows a bird view based on a Velodyne HDL64 point cloud (Geiger et al. \cite{Geiger:2012:WRA:2354409.2354978}) such as the predicted  objects. The lower one outlines the re-projection of the 3D boxes into image space. Note: Complex-YOLO needs no camera image as input, it is Lidar based only.}
\label{fig:Overview}
\end{figure}

In general, object detection and classification based on deep learning is a well known task and widely established for 2D bounding box regression on images \cite{DBLP:journals/corr/RedmonDGF15} \cite{DBLP:journals/corr/RedmonF16} \cite{DBLP:journals/corr/LiuAESR15} \cite{DBLP:journals/corr/RenHG015} \cite{DBLP:journals/corr/CaiFFV16} \cite{DBLP:journals/corr/RenCLSPYTX17} \cite{cvpr16chen} \cite{DBLP:journals/corr/GirshickDDM13} \cite{DBLP:journals/corr/HeZRS15} \cite{DBLP:journals/corr/ChenKZMFU16}. Research focus was mainly a trade-off between accuracy and efficiency. In regard to automated driving, efficiency is much more important. Therefore, the best object detectors are using region proposal networks (RPN) \cite{DBLP:journals/corr/abs-1711-06396} \cite{DBLP:journals/corr/Girshick15} \cite{DBLP:journals/corr/RenHG015} or a similar grid based RPN-approach \cite{DBLP:journals/corr/RedmonF16}. Those networks are extremely efficient, accurate and even capable of running on a dedicated hardware or embedded devices. Object detections on point clouds are still rarely, but more and more important. Those applications need to be capable of predicting 3D bounding boxes. Currently, there exist mainly three different approaches using deep learning \cite{DBLP:journals/corr/abs-1711-06396}: 
\begin{enumerate}
\item Direct point cloud processing using Multi-Layer-Perceptrons \cite{DBLP:journals/corr/abs-1711-08488} \cite{DBLP:journals/corr/QiSMG16} \cite{DBLP:journals/corr/QiYSG17} \cite{1801.07791} \cite{1801.07829}
\item Translation of Point-Clouds into voxels or image stacks by using Convolutional Neural Networks (CNN) \cite{DBLP:journals/corr/ChenMWLX16} \cite{DBLP:journals/corr/abs-1711-06396} \cite{DBLP:journals/corr/EngelckeRWTP16} \cite{Wang-RSS-15} \cite{DBLP:journals/corr/LiZX16} \cite{DBLP:journals/corr/Li16p} \cite{xiang_cvpr15} \cite{DBLP:journals/corr/WuSKTX14}
\item Combined fusion approaches \cite{DBLP:journals/corr/ChenMWLX16} \cite{ku2017joint}
\end{enumerate}

\subsection{Related Work}

Recently, Frustum-based Networks \cite{DBLP:journals/corr/abs-1711-08488} have shown high performance on the KITTI Benchmark suite. The model is ranked\footnote{The ranking refers to the time of submission: 14th of march in 2018} on the second place either for 3D object detections, as for birds-eye-view detection based on cars, pedestrians and cyclists. This is the only approach, which directly deals with the point cloud using Point-Net \cite{DBLP:journals/corr/QiSMG16} without using CNNs on Lidar data and voxel creation. However, it needs a pre-processing and therefore it has to use the camera sensor as well. Based on another CNN dealing with the calibrated camera image, it uses those detections to minimize the global point cloud to frustum-based reduced point cloud. This approach has two drawbacks: i). The models accuracy strongly depends on the camera image and its associated CNN. Hence, it is not possible to apply the approach to Lidar data only; ii). The overall pipeline has to run two deep learning approaches consecutive, which ends up in higher inference time with lower efficiency. The referenced  model runs with a too low frame-rate at approximately 7fps on a NVIDIA GTX 1080i GPU \cite{Geiger:2012:WRA:2354409.2354978}.  

In contrast, Zhou et al. \cite{DBLP:journals/corr/abs-1711-06396} proposed a model that operates only on Lidar data. In regard to that, it is the best ranked model on KITTI for 3D and birds-eye-view detections using Lidar data only. The basic idea is an end-to-end learning that operates on grid cells without using hand crafted features. Grid cell inside features are learned during training using a Pointnet approach \cite{DBLP:journals/corr/QiSMG16}. On top builds up a CNN that predicts the 3D bounding boxes. Despite the high accuracy, the model ends up in a low inference time of 4fps on a TitanX GPU \cite{DBLP:journals/corr/abs-1711-06396}.

Another highly ranked approach is reported by Chen et al. \cite{DBLP:journals/corr/abs-1711-08488}. The basic idea is the projection of Lidar point clouds into voxel based RGB-maps using handcrafted features, like points density, maximum height and a representative point intensity \cite{DBLP:journals/corr/Li16p}. To achieve highly accurate results, they use a multi-view approach based on a Lidar birds-eye-view map, a Lidar based front-view map and a camera based front-view image. This fusion ends up in a high processing time resulting in only 4fps on a NVIDIA GTX 1080i GPU. Another drawback is the need of the secondary sensor input (camera).

\subsection{Contribution}

To our surprise, no one is achieving real-time efficiency in terms of autonomous driving so far. Hence, we introduce the first slim and accurate model that is capable of running faster than 50fps on a NVIDIA TitanX GPU. We use the multi-view idea (MV3D) \cite{DBLP:journals/corr/abs-1711-08488} for point cloud pre-processing and feature extraction. However, we neglect the multi-view fusion and generate one single birds-eye-view RGB-map (see Fig. \ref{fig:Overview}) that is based on Lidar only, to ensure efficiency. On top, we present Complex-YOLO, a 3D version of YOLOv2, which is one of the fastest state-of-the-art image object detectors \cite{DBLP:journals/corr/RedmonF16}. Complex-YOLO is supported by our specific E-RPN that estimates the orientation of objects coded by an imaginary and real part for each box. The idea is to have a closed mathematical space without singularities for accurate angle generalization. Our model is capable to predict exact 3D boxes with localization and an exact heading of the objects in real-time, even if the object is based on a few points (e.g. pedestrians). Therefore, we designed special anchor-boxes. Further, it is capable to predict all eight KITTI classes by using only Lidar input data. We evaluated our model on the KITTI benchmark suite. In terms of accuracy, we achieved on par results for cars, pedestrians and cyclists, in terms of efficiency we outperform current leaders by minimum factor of 5. The main contributions of this paper are:

\begin{enumerate}
\item This work introduces Complex-YOLO by using a new E-RPN for reliable angle regression for 3D box estimation.
\item We present real-time performance with high accuracy evaluated on the KITTI benchmark suite by being more than five times faster than the current leading models.
\item We estimate an exact heading of each 3D box supported by the E-RPN that enables the prediction of the trajectory of surrounding objects.
\item Compared to other Lidar based methods (e.g. \cite{DBLP:journals/corr/abs-1711-06396}) our model efficiently estimates all classes simultaneously in one forward path.
\end{enumerate}

\section{Complex-YOLO}

This section describes the grid based pre-processing of the point clouds, the specific network architecture, the derived loss function for training and our efficiency design to ensure real-time performance.

\subsection{Point Cloud Preprocessing} \label{sec:pcl-pr}
\begin{figure}[!t]
\centering
\includegraphics[width=\textwidth]{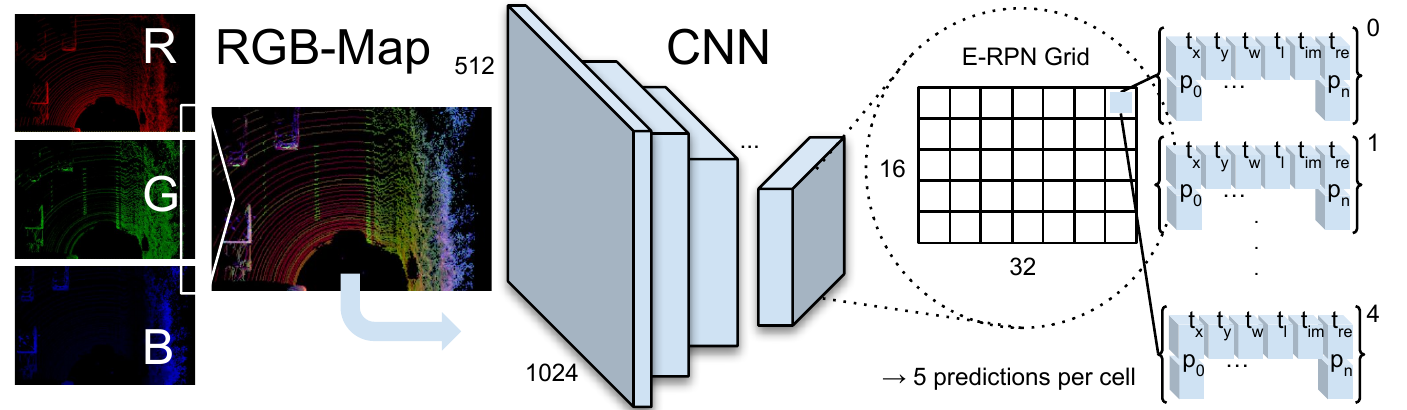}
\caption{\textbf{Complex-YOLO Pipeline.} We present a slim pipeline for fast and accurate 3D box estimations on point clouds. The RGB-map is fed into the CNN (see Tab. \ref{tab:arch}). The E-RPN grid runs simultaneously on the last feature map and predicts five boxes per grid cell. Each box prediction is composed by the regression parameters $t$ (see Fig. \ref{fig:OrientedRegion}) and object scores $p$ with a general probability $p_0$ and $n$ class scores $p_1 ... p_n$. }
\label{fig:Yolov2_Architecture}
\end{figure}

The 3D point cloud of a single frame, acquired by Velodyne HDL64 laser scanner \cite{Geiger:2012:WRA:2354409.2354978}, is converted into a single birds-eye-view RGB-map, covering an area of 80m x 40m (see Fig.\ref{fig:DatasetHeatmap}) directly in front of the origin of the sensor. Inspired by Chen et al. (MV3D) \cite{DBLP:journals/corr/abs-1711-08488}, the RGB-map is encoded by height, intensity and density. The size of the grid map is defined with $n=1024$ and $m=512$. Therefore, we projected and discretized the 3D point clouds into a 2D grid with resolution of about $g = 8cm$. Compared to MV3D, we slightly decreased the cell size to achieve less quantization errors, accompanied with higher input resolution. Due to efficiency and performance reasons, we are using only one instead of multiple height maps. Consequently, all three feature channels ($z_r, z_g, z_b$ with $z_{r,g,b} \in \mathbb{R}^{m \times n}$) are calculated for the whole point cloud $\mathcal{P} \in \mathbb{R}^3$ inside the covering area  $\Omega$. We consider the Velodyne within the origin of $\mathcal{P}_\Omega$ and define: 
\begin{equation}
\mathcal{P}_{\Omega} = \lbrace\mathcal{P}= [x,y,z]^T | x\in[0,40m], y \in[-40m,40m], z \in[-2m, 1.25m] \rbrace
\end{equation}
We choose $z \in[-2m, 1.25m]$, considering the Lidar z position of 1.73m \cite{Geiger:2012:WRA:2354409.2354978}, to cover an area above the ground to about 3m height, expecting trucks as highest objects. With the aid of the calibration \cite{Geiger:2012:WRA:2354409.2354978}, we define a mapping function $\mathcal{S}_j = f_{\mathcal{P}\mathcal{S}}(\mathcal{P}_{\Omega i}, g)$ with $ \mathcal{S} \in \mathbb{R}^{m\times n}$ mapping each point with index $i$ into a specific grid cell $\mathcal{S}_j$ of our RGB-map. A set describes all points mapped into a specific grid cell: 
\begin{equation}
\mathcal{P}_{\Omega i\rightarrow j} =\lbrace \mathcal{P}_{\Omega i} =[x,y,z]^T | \mathcal{S}_j = f_{\mathcal{P}\mathcal{S}}(\mathcal{P}_{\Omega i}, g)\rbrace
\end{equation}
Hence, we can calculate the channel of each pixel, considering the Velodyne intensity as $I(\mathcal{P}_\Omega)$:
\begin{align}
\begin{split}
z_g(\mathcal{S}_j) &= \max(\mathcal{P}_{\Omega i\rightarrow j} \cdot [0,0,1]^T) \\
z_b(\mathcal{S}_j) &= \max(I(\mathcal{P}_{\Omega i\rightarrow j}))\\
\label{point_cloud}
z_r(\mathcal{S}_j) &= \min\left(1.0,\log(N+1) / 64\right) \quad N=|\mathcal{P}_{\Omega i\rightarrow j}|
\end{split}
\end{align}
Here, $N$ describes the number of points mapped from  $\mathcal{P}_{\Omega i}$ to $\mathcal{S}_j$, and $g$ is the parameter for the grid cell size. Hence, $z_g$ encodes the maximum height, $z_b$ the maximum intensity and $z_r$ the normalized density of all points mapped into $\mathcal{S}_j$ (see Fig. \ref{fig:Yolov2_Architecture}).

\subsection{Architecture}

The Complex-YOLO network takes a birds-eye-view RGB-map (see section \ref{sec:pcl-pr}) as input. It uses a simplified YOLOv2 \cite{DBLP:journals/corr/RedmonF16} CNN architecture (see Tab. \ref{Tab:ComplexYoloDesign}), extended by a complex angle regression and E-RPN, to detect accurate multi-class oriented 3D objects while still operating in real-time.

\subsubsection{Euler-Region-Proposal.}
Our E-RPN parses the 3D position $b_{x,y}$, object dimensions (width $b_{w}$ and length $b_{l}$) as well as a probability $p_{0}$, class scores $p_{1}...p_{n}$ and finally its orientation $b_{\phi}$ from the incoming feature map. In order to get proper orientation, we have modified the commonly used Grid-RPN approach, by adding a complex angle $arg(|z|e^{ib_{\phi}})$ to it:
\begin{align}
\begin{split}
b_x &= \sigma(t_x) + c_x \\
b_y &= \sigma(t_y) + c_y \\
b_w &= p_w e^{t_w} \\
b_l &= p_l e^{t_l} \\
b_\phi &= arg(|z|e^{ib_{\phi}}) = \arctan_2(t_{Im},t_{Re})
\end{split}
\end{align}
With the help of this extension the E-RPN estimates accurate object orientations based on an imaginary and real fraction directly embedded into the network. For each grid cell (32x16 see Tab. \ref{Tab:ComplexYoloDesign}) we predict five objects including a probability score and class scores resulting in 75 features each, visualized in Fig. \ref{fig:Yolov2_Architecture}.

\begin{wraptable}[26]{r}{5.5cm}
\vspace{-1.15cm}
\centering
\caption{\textbf{Complex-YOLO Design.} Our ﬁnal model has 18 convolutional and 5 maxpool layers, as well as 3 intermediate layers for feature reorganization respectively.}
\label{Tab:ComplexYoloDesign}
\vspace{14pt}
\resizebox{0.45\textwidth}{!}{%
\begin{tabular}{c|c|c|c|c}
\textbf{layer}& \textbf{filters} & \textbf{size} & \textbf{input} & \textbf{output} \\ \hline
conv           & 24               & 3x3/1         & 1024x512x3     & 1024x512x24     \\
max            &                  & 2x2/2         & 1024x512x24    & 512x256x24      \\
conv           & 48               & 3x3/1         & 512x256x24     & 512x256x48      \\
max            &                  & 2x2/2         & 512x256x48     & 256x128x48      \\
conv           & 64               & 3x3/1         & 256x128x48     & 256x128x64      \\
conv           & 32               & 1x1/1         & 256x128x64     & 256x128x32      \\
conv           & 64               & 3x3/1         & 256x128x32     & 256x128x64      \\
max            &                  & 2x2/2         & 256x128x64     & 128x64x64       \\
conv           & 128              & 3x3/1         & 128x64x64      & 128x64x128      \\
conv           & 64               & 3x3/1         & 128x64x128     & 128x64x64       \\
conv           & 128              & 3x3/1         & 128x64x64      & 128x64x128      \\
max            &                  & 2x2/2         & 128x64x128     & 64x32x128       \\
conv           & 256              & 3x3/1         & 64x32x128      & 64x32x256       \\
conv           & 256              & 1x1/1         & 64x32x256      & 64x32x256       \\
conv           & 512              & 3x3/1         & 64x32x256      & 64x32x512       \\
max            &                  & 2x2/2         & 64x32x512      & 32x16x512       \\
conv           & 512              & 3x3/1         & 32x16x512      & 32x16x512       \\
conv           & 512              & 1x1/1         & 32x16x512      & 32x16x512       \\
conv           & 1024             & 3x3/1         & 32x16x512      & 32x16x1024      \\
conv           & 1024             & 3x3/1         & 32x16x1024     & 32x16x1024      \\
conv           & 1024             & 3x3/1         & 32x16x1024     & 32x16x1024      \\
route          & 12               &               &      &        \\
reorg          &                  &  /2           & 64x32x256      & 32x16x1024    \\
route          & 22 20            &               &      &    \\
conv           & 1024             & 3x3/1         & 32x16x2048    & 32x16x1024 \\   
conv           & 75               & 1x1/1         & 32x16x1024    & 32x16x75   \\ \hline
\textbf{E-RPN}&                  &               & \textbf{32x16x75}      &
\end{tabular}}
\label{tab:arch}
\end{wraptable}

\subsubsection{Anchor Box Design.}
The YOLOv2 object detector \cite{DBLP:journals/corr/RedmonF16} predicts five boxes per grid cell. All were initialized with beneficial priors, i.e. anchor boxes, for better convergence during training. Due to the angle regression, the degrees of freedom, i.e. the number of possible priors increased, but we did not enlarge the number of predictions for efficiency reasons. Hence, we defined only three different sizes and two angle directions as priors, based on the distribution of boxes within the KITTI dataset: i) vehicle size (heading up); ii) vehicle size (heading down); iii) cyclist size (heading up); iv) cyclist size (heading down); v) pedestrian size (heading left).

\subsubsection{Complex Angle Regression.}
The orientation angle for each object $b_{\phi}$ can be computed from the responsible regression parameters $t_{im}$ and $t_{re}$, which correspond to the phase of a complex number, similar to \cite{Beyer2015}. The angle is given simply by using $\arctan_2(t_{im},t_{re})$. On one hand, this avoids singularities, on the other hand this results in a closed mathematical space, which consequently has an advantageous impact on generalization of the model. We can link our regression parameters directly into the loss function (\ref{equ:loss}).

\begin{figure}[!t]
\centering
\includegraphics[width=0.45\textwidth]{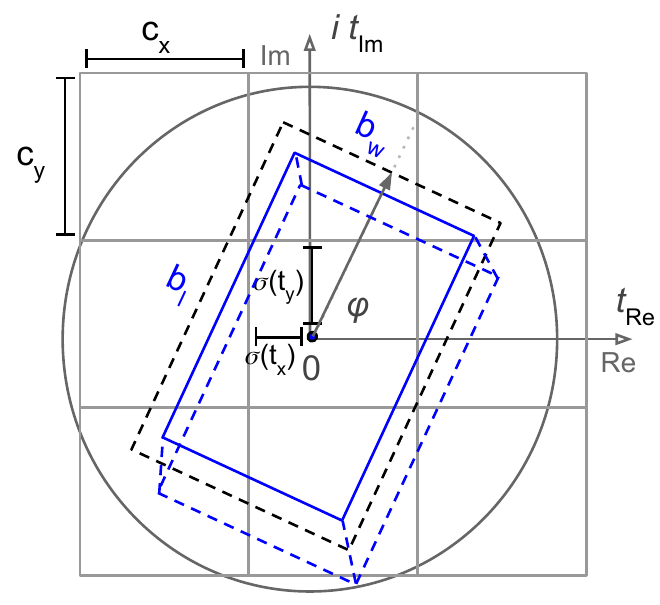}
\caption{\textbf{3D Bounding box regression.} We predict oriented 3D bounding boxes based on the regression parameters shown in YOLOv2 \cite{DBLP:journals/corr/RedmonF16}, as well as a complex angle for box orientation. The transition from 2D to 3D is done by a predefined height based on each class.}
\label{fig:OrientedRegion}
\end{figure}

\subsection{Loss Function}
Our network optimization loss function $\mathcal{L}$ is based on the the concepts from YOLO \cite{DBLP:journals/corr/RedmonDGF15} and YOLOv2 \cite{DBLP:journals/corr/RedmonF16}, who defined $\mathcal{L}_{\text{Yolo}}$ as the sum of squared errors using the introduced multi-part loss. We extend this approach by an Euler regression part $\mathcal{L}_{\text{Euler}}$ to get use of the complex numbers, which have a closed mathematical space for angle comparisons. This neglect singularities, which are common for single angle estimations:
\begin{equation}
\mathcal{L} = \mathcal{L}_{\text{Yolo}} + \mathcal{L}_{\text{Euler}}
\end{equation}
The Euler regression part of the loss function is defined with the aid of the Euler-Region-Proposal (see Fig. \ref{fig:OrientedRegion}). Assuming that the difference between the complex numbers of prediction and ground truth, i.e. $|z|e^{\mathrm{i} b_\phi}$ and $|\hat{z}|e^{\mathrm{i} \hat{b}_\phi}$ is always located on the unit circle with $|z|=1$ and $|\hat{z}|=1$, we minimize the absolute value of the squared error to get a real loss:
\begin{align}
\mathcal{L}_{\text{Euler}} &= \lambda_{coord}\sum_{i=0}^{S^2}\sum_{j=0}^B\mathds{1}_{ij}^{obj}\left|(e^{\mathrm{i} b_\phi} - e^{\mathrm{i} \hat{b}_\phi})^2\right| \\
&= \lambda_{coord}\sum_{i=0}^{S^2}\sum_{j=0}^B\mathds{1}_{ij}^{obj} \left[ (t_{im} - \hat{t}_{im})^2 + ( t_{re}-\hat{t}_{re})^2 \right]
\label{equ:loss}
\end{align}
Where $\lambda_{coord}$ is a scaling factor to ensure stable convergence in early phases and $\mathds{1}_{ij}^{obj}$ denotes that the $j$th bounding box predictor in cell $i$ has highest intersection over union (IoU) compared to ground truth for that prediction. 
Furthermore the comparison between the predicted box $P_j$ and ground truth $G$ with IoU $\frac{P_j \cap G}{P_j \cup G}$, where ${P_j \cap G = \{x : x \in P_j \land x \in G\}}$, ${P_j \cup G \{x : x \in P_j \lor x \in G\}}$ is adjusted to handle rotated boxes as well. This is realized by the theory of intersection of two 2D polygon geometries and union respectively, generated from the corresponding box parameters $b_x$, $b_y$, $b_w$, $b_l$ and $b_{\phi}$. 

\subsection{Efficiency Design}
The main advantage of the used network design is the prediction of all bounding boxes in one inference pass. The E-RPN is part of the network and uses the output of the last convolutional layer to predict all bounding boxes. Hence, we only have one network, which can be trained in an end-to-end manner without specific training approaches. Due to this, our model has a lower runtime than other models that generate region proposals in a sliding window manner \cite{DBLP:journals/corr/Girshick15} with  prediction of offsets and the class for every proposal (e.g. Faster R-CNN~\cite{DBLP:journals/corr/RenHG015}). In Fig. \ref{fig:mAPFPS} we compare our architecture with some of the leading models on the KITTI benchmark. Our approach achieves a way higher frame rate while still keeping a comparable mAP (mean Average Precision). The frame rates were directly taken from the respective papers and all were tested on a Titan X or Titan Xp. We tested our model on a Titan X and an NVIDIA TX2 board to emphasize the real-time capability (see Fig. \ref{fig:mAPFPS}).

\section{Training \& Experiments}
We evaluated Complex-YOLO on the challenging KITTI object detection benchmark \cite{Geiger:2012:WRA:2354409.2354978}, which is divided into three subcategories 2D, 3D and birds-eye-view object detection for \textit{Cars}, \textit{Pedestrians} and \textit{Cyclists}. Each class is evaluated based on three difficulty levels \textit{easy}, \textit{moderate} and \textit{hard} considering the object size, distance, occlusion and truncation. This public dataset provides 7,481 samples for training including annotated ground truth and 7,518 test samples with point clouds taken from a Velodyne laser scanner, where annotation data is private. Note that we focused on birds-eye-view and do not ran the 2D object detection benchmark, since our input is Lidar based only.

\subsection{Training Details}
We trained our model from scratch via stochastic gradient descent with a weight decay of 0.0005 and momentum 0.9. Our implementation is based on modified version of the Darknet neural network framework \cite{darknet13}. First, we applied our pre-processing (see Section \ref{sec:pcl-pr}) to generate the birds-eye-view RGB-maps from Velodyne samples. Following the principles from \cite{DBLP:journals/corr/ChenMWLX16} \cite{DBLP:journals/corr/abs-1711-06396} \cite{Chen2015NIPS}, we subdivided the training set with public available ground truth, but used ratios of 85\% for training and 15\% for validation, because we trained from scratch and aimed for a model that is capable of multi-class predictions. In contrast, e.g. VoxelNet \cite{DBLP:journals/corr/abs-1711-06396} modified and optimized the model for different classes. We suffered from the available ground truth data, because it was intended for camera detections first. The class distribution with more than 75\% \textit{Car}, less than 4\% \textit{Cyclist} and less than 15\% \textit{Pedestrian} is disadvantageous. Also, more than 90\% of all the annotated objects are facing the car direction, facing towards the recording car or having similar orientations. On top, Fig. \ref{fig:DatasetHeatmap} shows a 2D histogram for spatial object locations from birds-eye-view perspective, where dense points indicate more objects at exactly this position. This inherits two blind spot for birds-eye-view map. Nevertheless we saw surprising good results for the validation set and other recorded unlabeled KITTI sequences covering several use case scenarios, like urban, highway or inner city. 

For the first epochs, we started with a small learning rate to ensure convergence. After some epochs, we scaled the learning rate up and continued to gradually decrease it for up to 1,000 epochs. Due to the fine grained requirements, when using a birds-eye-view approach, slight changes in predicted features will have a strong impact on resulting box predictions. We used batch normalization for regularization and a linear activation $f(x) = x$ for the last layer of our CNN, apart from that the leaky rectified linear activation:

\begin{figure}[!t]
\centering
	\includegraphics[width=\textwidth]{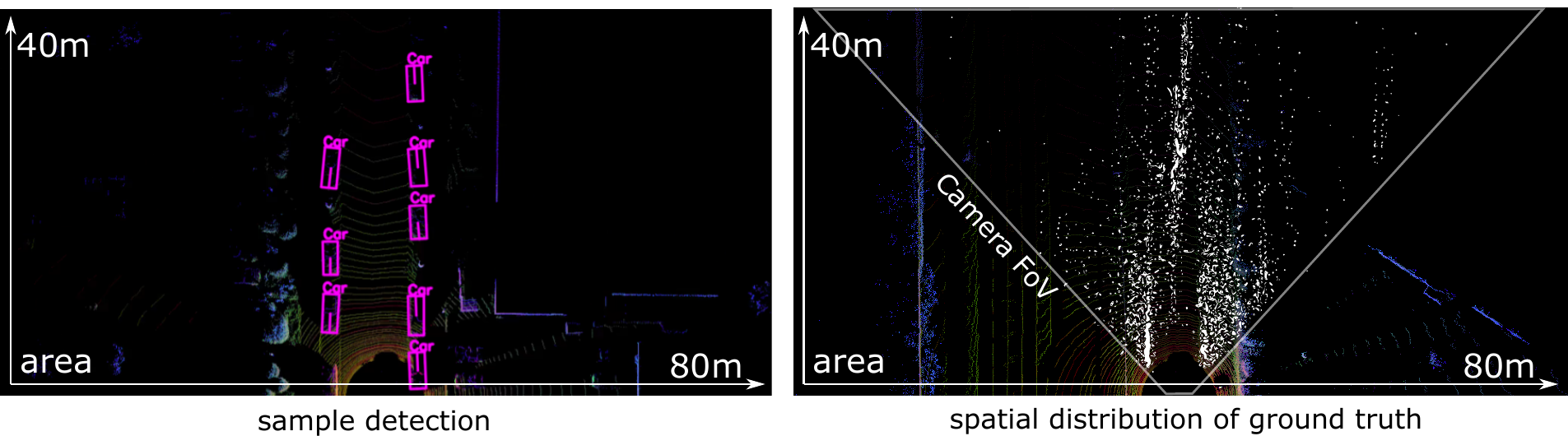}
	\caption{\textbf{Spatial ground truth distribution.} The figure outlines the size of the birds-eye-view area with a sample detection on the left. The right shows a 2D spatial histogram of annotated boxes in \cite{Geiger:2012:WRA:2354409.2354978}. The distribution outlines the horizontal field of view of the camera used for annotation and the inherited blind spots in our map.}
	\label{fig:DatasetHeatmap}
\end{figure}

\begin{figure}[!t]
\centering
	\includegraphics[width=0.65\textwidth]{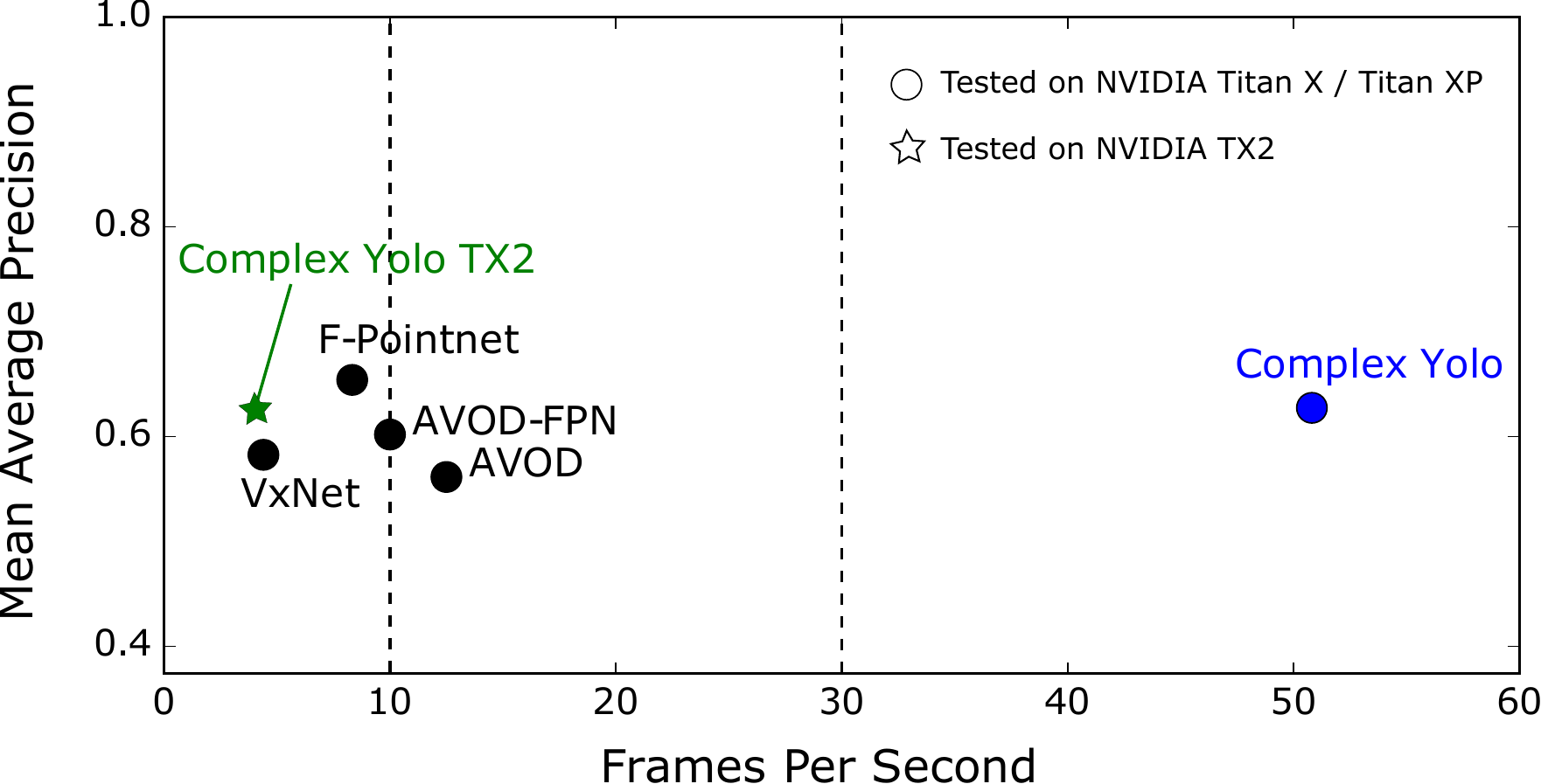}
	\caption{\textbf{Performance comparison.} This plot shows the mAP in relation to the runtime (fps). All models were tested on a Nvidia Titan X or Titan Xp. Complex-Yolo achieves accurate results by being five times faster than the most effective competitor on the KITTI benchmark \cite{Geiger:2012:WRA:2354409.2354978}. We compared to the five leading models and measured our network on a dedicated embedded platform (TX2) with reasonable efficiency (4fps) as well. Complex-Yolo is the first model for real-time 3D object detection.}
	\label{fig:mAPFPS}
\end{figure}

\begin{equation}
f(x) = \left\{\begin{array}{ll} x, & x > 0\\ 0.1x, & \text{otherwise}\end{array} \right.
\end{equation}

\subsection{Evaluation on KITTI}
We have adapted our experimental setup and follow the official KITTI evaluation protocol, where the IoU thresholds are 0.7 for class \textit{Car} and 0.5 for class \textit{Pedestrian} and \textit{Cyclist}. Detections that are not visible on the image plane are filtered, because the ground truth is only available for objects that also appear on the image plane of the camera recording \cite{Geiger:2012:WRA:2354409.2354978} (see Fig. \ref{fig:DatasetHeatmap}. We used the average precision (AP) metric to compare the results. Note, that we ignore a small number of objects that are outside our birds-eye-view map boundaries with more than 40m to the front, to keep the input dimensions as small as possible for efficiency reasons.

\subsubsection{Birds-Eye-View.} Our evaluation results for the birds-eye-view detection are presented in Tab. \ref{Tab:PerformanceComparisonBV}. This benchmark uses bounding box overlap for comparison. For a better overview and to rank the results, similar current leading methods are listed as well, but performing on the official KITTI test set. Complex-YOLO consistently outperforms all competitors in terms of runtime and efficiency, while still manages to achieve comparable accuracy. With about 0.02s runtime on a Titan X GPU, we are 5 times faster than AVOD \cite{ku2017joint}, considering their usage of a more powerful GPU (Titan Xp). Compared to VoxelNet \cite{DBLP:journals/corr/abs-1711-06396}, which is also Lidar based only, we are more than 10 times faster and MV3D \cite{DBLP:journals/corr/ChenMWLX16}, the slowest competitor, takes 18 times as long.

\begin{table}
\begin{center}
\caption{\textbf{Performance comparison for birds-eye-view detection.} APs (in \%) for our experimental setup compared to current leading methods. Note that our method is validated on our splitted validation dataset, whereas all others are validated on the official KITTI test set.}
\label{Tab:PerformanceComparisonBV}
\resizebox{12.2cm}{!}{%
\renewcommand{\arraystretch}{1.4}
\begin{tabular}{|>{\centering\arraybackslash}m{2.2cm}|>{\centering\arraybackslash}m{2.0cm}|>{\centering\arraybackslash}m{1cm}||>{\centering\arraybackslash}m{1cm}|>{\centering\arraybackslash}m{1cm}|>{\centering\arraybackslash}m{1cm}||>{\centering\arraybackslash}m{1cm}|>{\centering\arraybackslash}m{1cm}|>{\centering\arraybackslash}m{1cm}||>{\centering\arraybackslash}m{1cm}|>{\centering\arraybackslash}m{1cm}|>{\centering\arraybackslash}m{1cm}||}
\hline 
\multirow{2}{*}{\large{\textbf{Method}}} & \multirow{2}{*}{\large{\textbf{Modality}}} & \multirow{2}{*}{\large{\textbf{FPS}}} & \multicolumn{3}{c||}{\large{\textbf{Car}}} & \multicolumn{3}{c||}{\large{\textbf{Pedestrian}}} & \multicolumn{3}{c||}{\large{\textbf{Cyclist}}} \\ 
 & & & \textbf{Easy} & \textbf{Mod.} & \textbf{Hard} & \textbf{Easy} & \textbf{Mod.} & \textbf{Hard} & \textbf{Easy} & \textbf{Mod.} & \textbf{Hard} \\
\hline
\hline
MV3D \cite{DBLP:journals/corr/ChenMWLX16} & Lidar\texttt{+}Mono & 2.8 & 86.02 & 76.90 & 68.49 & - & - & - & - & - & - \\
\hline
F-PointNet \cite{DBLP:journals/corr/abs-1711-08488} & Lidar\texttt{+}Mono & 5.9 & 88.70 & 84.00 & 75.33 & \textbf{58.09} & \textbf{50.22} & \textbf{47.20} & \textbf{75.38} & 61.96 & 54.68 \\
\hline
AVOD \cite{ku2017joint} & Lidar\texttt{+}Mono & 12.5 & 86.80 & \textbf{85.44} & 77.73 & 42.51 & 35.24 & 33.97 & 63.66 & 47.74 & 46.55 \\
\hline
AVOD-FPN \cite{ku2017joint} & Lidar\texttt{+}Mono & 10.0 & 88.53 & 83.79 & \textbf{77.90} & 50.66 & 44.75 & 40.83 & 62.39 & 52.02 & 47.87 \\
\hline
VoxelNet \cite{DBLP:journals/corr/abs-1711-06396} & Lidar & 4.3 & \textbf{89.35} & 79.26 & 77.39 & 46.13 & 40.74 & 38.11 & 66.70 & 54.76 & 50.55 \\
\hline
\hline
Complex-YOLO & Lidar & \textbf{50.4} & 85.89 & 77.40 & 77.33 & 46.08 & 45.90 & 44.20 & 72.37 & \textbf{63.36} & \textbf{60.27} \\
\hline 
\end{tabular}}
\end{center}
\end{table}

\subsubsection{3D Object Detection.} Tab. \ref{Tab:PerformanceComparison3D} shows our achieved results for the 3D bounding box overlap. Since we do not estimate the height information directly with regression, we ran this benchmark with a fixed spatial height location extracted from ground truth similar to MV3D \cite{DBLP:journals/corr/ChenMWLX16}. Additionally as mentioned, we simply injected a predefined height for every object based on its class, calculated from the mean over all ground truth objects per class. This reduces the precision for all classes, but it confirms the good results measured on the birds-eye-view benchmark.

\begin{table}
\begin{center}
\caption{\textbf{Performance comparison for 3D object detection.} APs (in \%) for our experimental setup compared to current leading methods. Note that our method is validated on our splitted validation dataset, whereas all others are validated on the official KITTI test set.}
\label{Tab:PerformanceComparison3D}
\resizebox{12.2cm}{!}{%
\renewcommand{\arraystretch}{1.4}
\begin{tabular}{|>{\centering\arraybackslash}m{2.2cm}|>{\centering\arraybackslash}m{2.0cm}|>{\centering\arraybackslash}m{1cm}||>{\centering\arraybackslash}m{1cm}|>{\centering\arraybackslash}m{1cm}|>{\centering\arraybackslash}m{1cm}||>{\centering\arraybackslash}m{1cm}|>{\centering\arraybackslash}m{1cm}|>{\centering\arraybackslash}m{1cm}||>{\centering\arraybackslash}m{1cm}|>{\centering\arraybackslash}m{1cm}|>{\centering\arraybackslash}m{1cm}||}
\hline 
\multirow{2}{*}{\large{\textbf{Method}}} & \multirow{2}{*}{\large{\textbf{Modality}}} & \multirow{2}{*}{\large{\textbf{FPS}}} & \multicolumn{3}{c||}{\large{\textbf{Car}}} & \multicolumn{3}{c||}{\large{\textbf{Pedestrian}}} & \multicolumn{3}{c||}{\large{\textbf{Cyclist}}} \\ 
 & & & \textbf{Easy} & \textbf{Mod.} & \textbf{Hard} & \textbf{Easy} & \textbf{Mod.} & \textbf{Hard} & \textbf{Easy} & \textbf{Mod.} & \textbf{Hard} \\
\hline
\hline
MV3D \cite{DBLP:journals/corr/ChenMWLX16} & Lidar\texttt{+}Mono & 2.8 & 71.09 & 62.35 & 55.12 & - & - & - & - & - & - \\
\hline
F-PointNet \cite{DBLP:journals/corr/abs-1711-08488} & Lidar\texttt{+}Mono & 5.9 & 81.20 & 70.39 & 62.19 & \textbf{51.21} & \textbf{44.89} & \textbf{40.23} & \textbf{71.96} & 56.77 & 50.39 \\
\hline
AVOD \cite{ku2017joint} & Lidar\texttt{+}Mono & 12.5 & 73.59 & 65.78 & 58.38 & 38.28 & 31.51 & 26.98 & 60.11 & 44.90 & 38.80 \\
\hline
AVOD-FPN \cite{ku2017joint} & Lidar\texttt{+}Mono & 10.0 & \textbf{81.94} & \textbf{71.88} & \textbf{66.38} & 46.35 & 39.00 & 36.58 & 59.97 & 46.12 & 42.36 \\
\hline
VoxelNet \cite{DBLP:journals/corr/abs-1711-06396} & Lidar & 4.3 & 77.47 & 65.11 & 57.73 & 39.48 & 33.69 & 31.51 & 61.22 & 48.36 & 44.37 \\
\hline
\hline
Complex-YOLO & Lidar & \textbf{50.4} & 67.72 & 64.00 & 63.01 & 41.79 & 39.70 & 35.92 & 68.17 & \textbf{58.32} & \textbf{54.30} \\
\hline 
\end{tabular}}
\end{center}
\end{table}

\section{Conclusion}
In this paper we present the first real-time efficient deep learning model for 3D object detection on Lidar based point clouds. We highlight our state of the art results in terms of accuracy (see Fig. \ref{fig:mAPFPS}) on the KITTI benchmark suite with an outstanding efficiency of more than 50 fps (NVIDIA Titan X). We do not need additional sensors, e.g. camera, like most of the leading approaches. This breakthrough is achieved by the introduction of the new E-RPN, an Euler regression approach for estimating orientations with the aid of the complex numbers. The closed mathematical space without singularities allows robust angle prediction. 

Our approach is able to detect objects of multiple classes (e.g. cars, vans, pedestrians, cyclists, trucks, tram, sitting pedestrians, misc) simultaneously in one forward path. This novelty enables deployment for real usage in self driving cars and clearly differentiates to other models. We show the real-time capability even on dedicated embedded platform NVIDIA TX2 (4 fps). In future work, it is planned to add height information to the regression, enabling a real independent 3D object detection in space, and to use tempo-spatial dependencies within point cloud pre-processing for a better class distinction and improved accuracy.

\section*{Acknowledgement}
First, we would like to thank our main employer Valeo, especially J\"org Schrepfer and Johannes Petzold, for giving us the possibility to do fundamental research. Additionally, we would like to thank our colleague Maximillian Jaritz for his important contribution on voxel generation. Last but not least, we would like to thank our academic partner the TU-Ilmenau for having a fruitful partnership.

\begin{figure}[!t]
\centering
\includegraphics[width=0.9\textwidth]{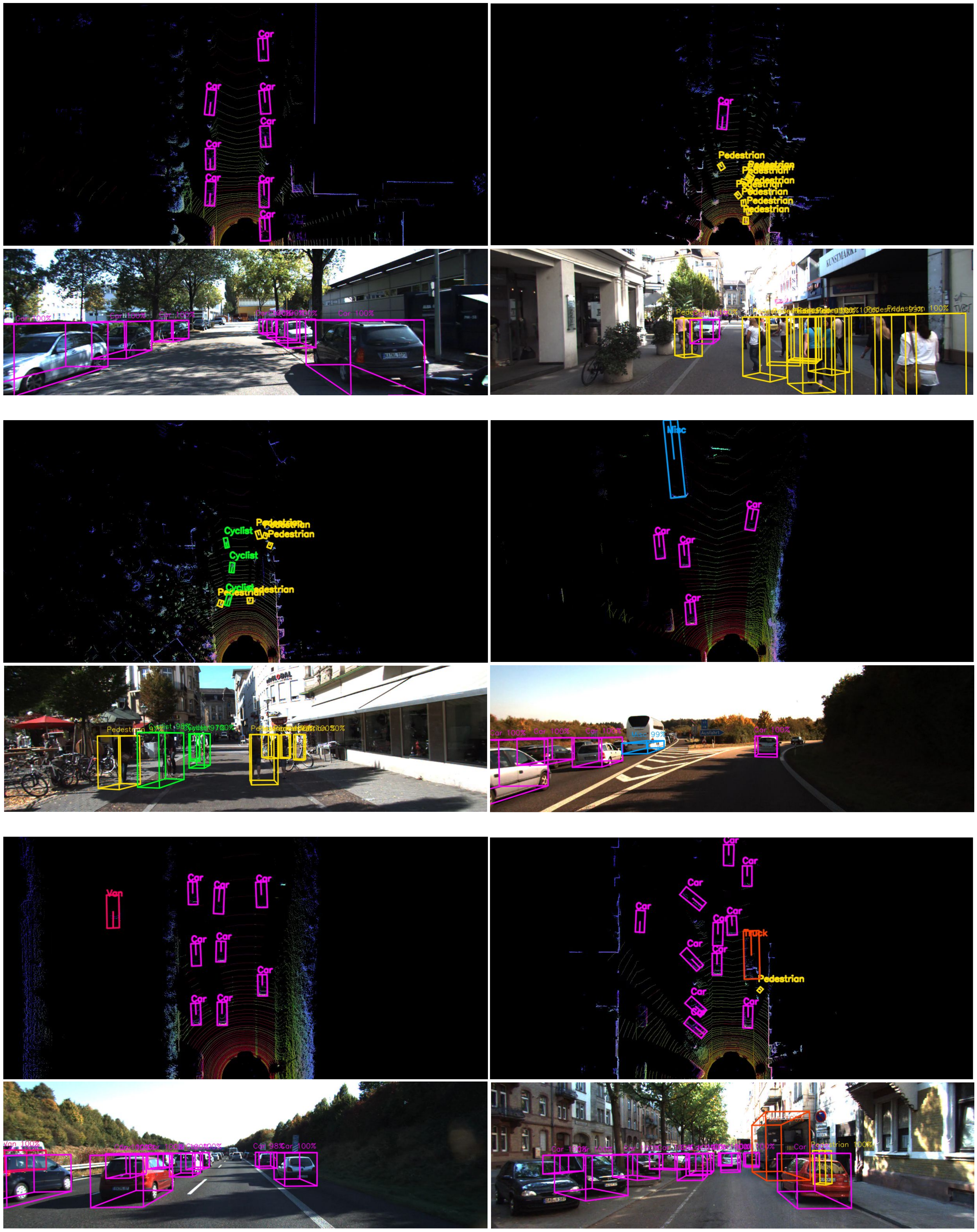}
\caption{\textbf{Visualization of Complex-YOLO results.} Note that predictions are exclusively based on birds-eye-view images generated from point clouds. The re-projection into camera space is for illustrative purposes only.}
\label{fig:QualitativeResults}
\end{figure}


\clearpage

\bibliographystyle{splncs}
\bibliography{egbib}
\end{document}